# QQJ: Quantifying Qualitative Judgment for Scalable and Human-Aligned Evaluation of Generative AI

Marjan Veysi
AI Lab, Arioobarzan Engineering Team, Shiraz, Iran, marjan@arioo.ir Pirooz Shamsinejadbabaki
Department of Computer Engineering and Information Technology, Shiraz University of Technology, Shiraz, Iran, p.shamsinejad@sutech.ac.ir Mohammad Zare
AI Lab, Arioobarzan Engineering Team, Shiraz, Iran, md.zare@sutech.ac.ir Mohammad Sabouri
Department of Informatics, Bioengineering, Robotics and Systems Engineering, University of Genoa, Genoa, Italy m.sabouri@shirazu.ac.ir

*Abstract*—The rapid progress of generative artificial intelligence has exposed fundamental limitations in existing evaluation methodologies, particularly for open-ended, creative, and human-facing tasks. Traditional automatic metrics rely on surface-level statistical similarity and often fail to reflect human perceptions of quality, while purely human evaluation, although reliable, is costly, subjective, and difficult to scale. Recent approaches using large language models as evaluators offer improved scalability but frequently lack explicit grounding in human-defined evaluation principles, leading to bias and inconsistency. In this paper, we introduce Quantifying Qualitative Judgment (QQJ), a scalable and human-centric evaluation framework that explicitly bridges the gap between human judgment and automated assessment. QQJ separates the definition of quality from its execution by anchoring evaluation in expert-designed, multi-dimensional rubrics and calibrating large language model evaluators to align with expert reasoning using a small, high-quality annotation set. This design enables consistent, interpretable, and scalable evaluation across diverse generative tasks and modalities. Extensive experiments on text and image generation demonstrate that QQJ achieves substantially stronger alignment with human judgment than traditional automatic metrics and unconstrained LLM-based evaluators. Moreover, QQJ exhibits improved stability across repeated evaluations and superior diagnostic capability in identifying critical failure modes such as hallucination and intent mismatch. These results indicate that structured qualitative judgment can be operationalized at scale without sacrificing interpretability or human alignment, positioning QQJ as a practical foundation for reliable evaluation of modern generative AI systems.

## I. Introduction

The rapid evolution of generative artificial intelligence has fundamentally reshaped the landscape of machine-generated content. Recent large language models and generative vision systems are capable of producing fluent text, high-fidelity images, and stylistically rich outputs that increasingly resemble human-created artifacts. These advances have enabled widespread adoption of generative models in domains such as content creation, decision support, education, and human–AI interaction. As the expressive capacity of generative systems continues to grow, the question of how to reliably and meaningfully evaluate their outputs has become both more critical and more complex [1], [2].

Figure 1 conceptually illustrates the central limitation of existing evaluation practices. While surface-level assessment may deem a generative output acceptable, deeper qualitative analysis reveals hidden issues that only structured evaluation frameworks such as QQJ are able to capture.

Despite the sophistication of modern generative models, evaluation methodologies have not kept pace with this progress. The dominant automatic metrics used in natural language generation and image synthesis remain largely rooted in surface-level statistical similarity, including n-gram overlap, likelihood estimates, or feature-space distances [3], [4]. These metrics were originally designed for constrained tasks with well-defined reference outputs and are poorly suited for open-ended generation, creative tasks, or scenarios where multiple valid outputs exist. As a result, high metric scores do not necessarily correspond to outputs that humans perceive as accurate, coherent, or aligned with their intent [5]. Furthermore, advancements in generative models, such as Multi-Fed GANs for privacy-preserving image colorization and Pix2Pix with attention mechanisms for image reconstruction, highlight the need for more reliable and nuanced evaluation techniques [6],
[7].

Human judgment remains the most reliable means of assessing qualitative dimensions such as factuality, coherence, intent alignment, and creativity. However, human evaluation is expensive, time-consuming, and difficult to scale, particularly for large benchmarks or iterative model development. Moreover, expert annotators frequently disagree on nuanced or creative tasks [8], [9].

Recent work has explored the use of large language models as automated evaluators, leveraging their reasoning capabilities to judge the outputs of generative systems [10], [11]. While promising, these approaches often lack explicit grounding in human-designed evaluation criteria and systematic calibration,

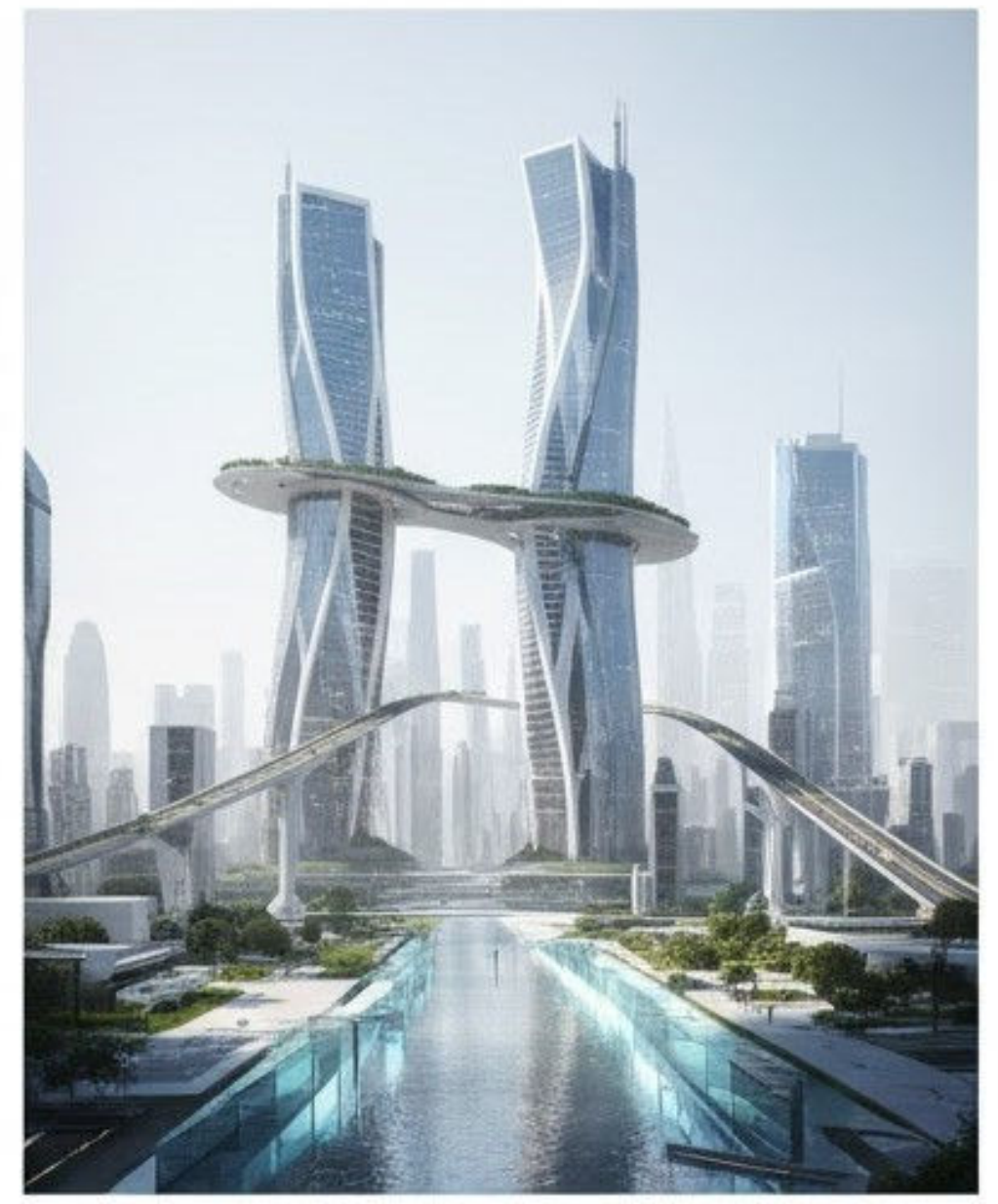

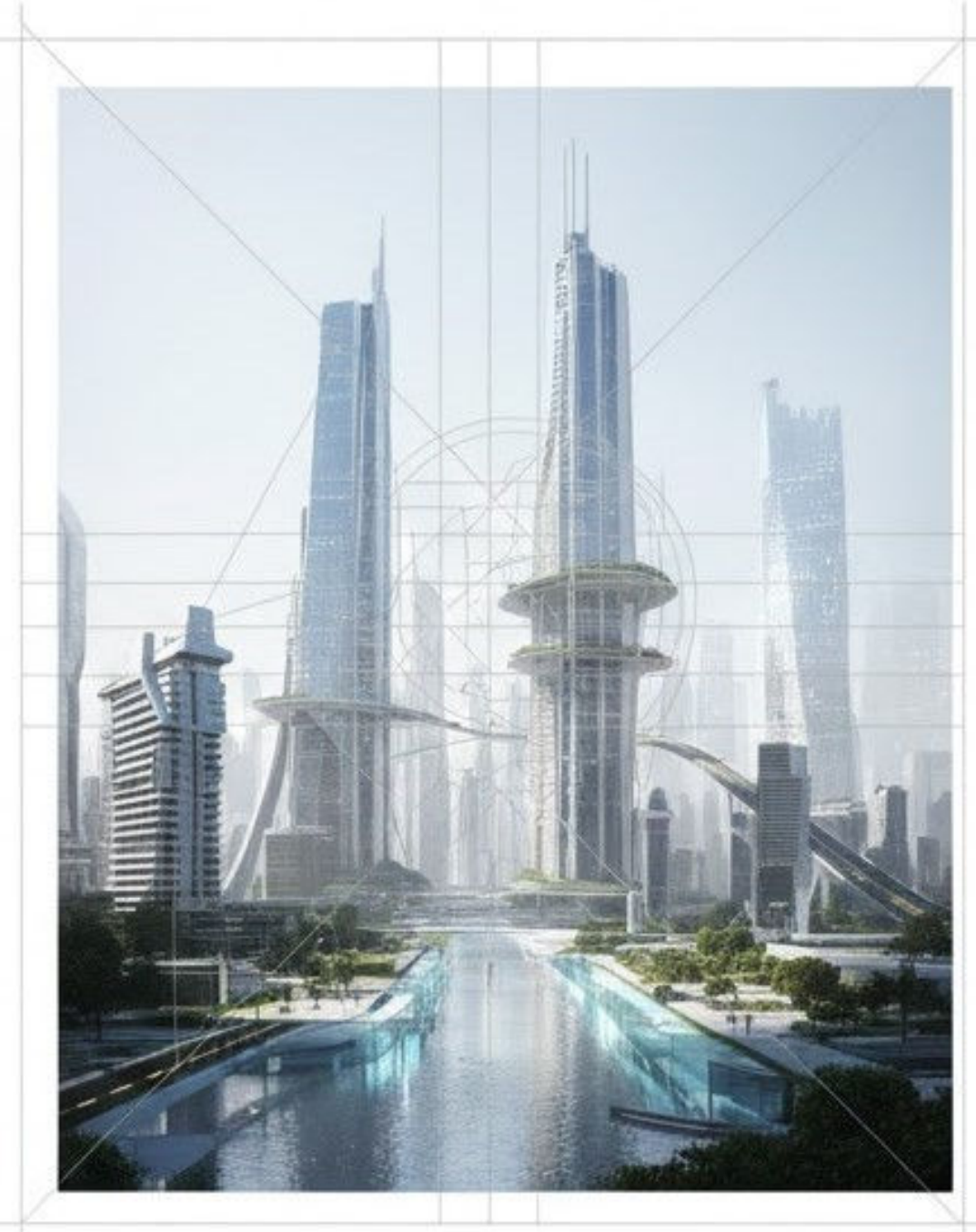

Fig. 1. Conceptual comparison of generative outputs without and with structured qualitative evaluation. While surface-level assessment may suggest comparable quality, structured evaluation frameworks such as QQJ enable more reliable and human-aligned assessment.

making them vulnerable to bias and inconsistency [12], [13].

As AI-driven generative systems have increasingly blurred the lines between human and machine creativity, discussions around their role as independent creators have gained significant attention, with studies exploring the intersection of computational creativity and AI's potential for artistic expression [14], [15].

Based on these observations, this paper introduces QQJ, a human-centric approach for evaluating generative AI systems at scale. The framework operationalizes qualitative assessment through structured criteria while leveraging large language models for consistent application. The remainder of this paper is organized as follows. Section II reviews related work. Section III presents the proposed framework. Section IV describes the experimental setup. Section V reports results. Section VI discusses limitations and implications, and Section VII concludes the paper.

## II. Related Work

As generative artificial intelligence systems have grown in expressive capacity, evaluation has emerged as a central challenge that directly affects model development, benchmarking practices, and deployment decisions. Early evaluation methodologies were designed for constrained generation tasks and relatively simple output distributions. However, the increasing complexity, openness, and subjectivity of modern generative outputs have exposed fundamental limitations in these traditional evaluation paradigms. Recent literature reflects a growing consensus that evaluation is no longer a secondary concern, but a core research problem in its own right.

### *A. Limitations of Traditional Automatic Metrics*

Traditional automatic metrics have long been favored due to their efficiency, reproducibility, and ease of comparison. Metrics such as BLEU, ROUGE, perplexity, Inception Score, and Frechet Inception Distance remain widely used across text´ and image generation benchmarks. Nevertheless, a substantial body of recent work demonstrates that these metrics fail to capture critical qualitative aspects of modern generative outputs, including semantic correctness, instruction adherence, long-range coherence, and creative intent.

Empirical studies show that optimizing for such metrics can encourage undesirable behaviors. In text generation, models may produce fluent but hallucinated responses that score highly under likelihood-based measures. In image generation, visually realistic outputs may achieve strong FID scores while violating compositional or stylistic constraints specified by the prompt. These findings suggest that traditional metrics primarily reflect surface-level statistical similarity rather than human-perceived quality, limiting their usefulness for evaluating contemporary generative systems.

### B. Human Evaluation as the Reference Standard

Human evaluation remains the most reliable mechanism for assessing generative quality, particularly for open-ended or subjective tasks. Human evaluators are capable of integrating multiple dimensions of quality, including factuality, coherence, usefulness, and stylistic appropriateness, into a holistic judgment. Consequently, many benchmarks and evaluation campaigns continue to rely on human judgments as the ultimate reference standard.

Despite its strengths, human evaluation suffers from well-documented limitations. Large-scale annotation is costly and time-consuming, and inter-annotator agreement is often limited for nuanced or creative tasks. Moreover, human judgments can vary depending on evaluator expertise, task framing, and rubric design. These constraints make purely human-driven evaluation impractical for large-scale benchmarking and iterative model development, motivating the search for scalable alternatives.

### C. Emergence of LLM-Based Evaluation

In response to the scalability limitations of human evaluation, recent research has explored the use of large language models as automated evaluators. This paradigm, commonly referred to as LLM-as-a-judge, leverages the reasoning and language understanding capabilities of LLMs to assess generative outputs across multiple dimensions. Prominent examples include MT-Bench for conversational agents and G-Eval for structured natural language generation assessment.

While these approaches often report encouraging correlations with human judgments, subsequent studies reveal important limitations. LLM-based evaluators are susceptible to systematic biases, including positional bias, verbosity bias, and agreement bias. Furthermore, without explicit grounding in human-defined evaluation criteria, such evaluators may rely on latent heuristics rather than principled reasoning. These concerns raise questions about the reliability, transparency, and reproducibility of unconstrained LLM-based evaluation methods.

### D. Structured and Rubric-Guided Evaluation

To address the shortcomings of both traditional metrics and unconstrained LLM judges, recent work emphasizes structured and rubric-guided evaluation frameworks. By explicitly defining quality dimensions and scoring criteria, rubric-based approaches improve transparency and interpretability. Benchmarks such as HELM, BIG-Bench, and GEM incorporate multi-dimensional evaluation protocols that go beyond single-score metrics and attempt to reflect a broader range of human evaluation criteria.

However, most existing rubric-based approaches still rely heavily on human annotation or task-specific evaluation pipelines, limiting scalability. In addition, alignment between automated evaluators and expert human judgments is often implicit or absent, reducing consistency across datasets and tasks. These gaps highlight the need for evaluation frameworks that combine explicit human grounding with scalable and consistent execution.

### E. Positioning with Respect to Prior Work

Taken together, prior work reveals a structural gap in generative model evaluation. Traditional automatic metrics provide scalability but lack semantic depth, while human evaluation offers nuanced judgment but does not scale. LLM-based evaluators occupy an intermediate position, yet often lack explicit alignment with human evaluation principles and rigorous calibration. Recent studies increasingly suggest that effective evaluation requires separating the definition of quality from its operationalization, allowing human expertise to guide scalable automated assessment.

### F. Comparative Summary of Prior Approaches

To contextualize the proposed approach within the broader literature, Table I summarizes representative evaluation methods and contrasts them across key dimensions relevant to generative AI evaluation.

## III. Methodology

This section presents the proposed methodology for Quantifying Qualitative Judgment (QQJ). The primary objective of QQJ is to enable scalable evaluation of generative AI outputs while preserving the structure, nuance, and interpretability of human qualitative judgment. In contrast to existing evaluation approaches that either rely on implicit heuristics or unconstrained model-based scoring, QQJ explicitly separates the definition of quality from its execution. This separation assigns complementary roles to human experts and large language models, aligning with recent calls for human-grounded yet scalable evaluation frameworks [5], [9], [24].

### A. Overview of the Evaluation Pipeline

Figure 2 illustrates the end-to-end evaluation pipeline of QQJ. The pipeline consists of four sequential stages: qualitative rubric construction, expert annotation and calibration set creation, evaluator alignment, and scalable automated evaluation. This staged design reflects emerging best practices in human-in-the-loop evaluation, where human expertise is used to define evaluation principles, and automation is applied only after these principles are made explicit and verifiable [8], [20].

### B. Qualitative Rubric Construction

The first stage of QQJ focuses on the explicit formalization of qualitative evaluation criteria. Rather than relying on task-specific metrics or latent notions of quality, QQJ employs a multi-dimensional rubric designed by human experts. Each rubric dimension corresponds to a distinct aspect of human judgment, such as factual fidelity, semantic coherence, intent alignment, or creative appropriateness. The use of structured rubrics is motivated by prior findings showing that explicit criteria improve both consistency and interpretability in human and automated evaluation [25], [26].

Let $\mathrm{R} = \{r_1, r_2, \ldots, r_K\}$ denote the set of qualitative dimensions defined in the rubric. Each dimension $r_k$ is associated with a semantic definition and a clearly specified scoring guideline, which may be ordinal or scalar depending on the task. The goal of this stage is not to directly compute a quality score, but to externalize human judgment into a structured and reproducible form that can later be operationalized.

### *C. Expert Annotation and Calibration Set*

Given the rubric $\mathrm{R}$, a small but carefully curated set of generative outputs is annotated by domain experts. This calibration set serves as the reference for aligning automated evaluation with human reasoning. In contrast to large-scale human evaluation campaigns, QQJ prioritizes annotation fidelity and inter-annotator reliability over volume, following evidence that high-quality expert annotations are more informative for evaluation alignment than large quantities of noisy labels [9], [24].

Formally, let $\mathcal{D}_c = \{(x_i, y_i)\}_{i=1}^{N}$ denote the calibration dataset, where $x_i$ represents a generative output and $y_i = (y_{i1}, \ldots, y_{iK})$ denotes the vector of expert-assigned scores for each rubric dimension. The size $N$ is intentionally limited to maintain feasibility while ensuring coverage of representative model behaviors, including both successful and failure cases such as hallucinations or intent violations [27].

TABLE I

COMPARISON OF REPRESENTATIVE EVALUATION APPROACHES FOR GENERATIVE AI

| Approach | Human-Grounded | Scalable | Multi-Dimensional | Calibrated | Modalities | Year |
|---|---|---|---|---|---|---|
| BLEU / ROUGE [16], [17] | No | Yes | No | No | Text | 2002–2004 |
| FID / IS [18], [19] | No | Yes | No | No | Image | 2016–2017 |
| GEM Benchmark [8] | Yes | No | Yes | No | Text | 2023 |
| HELM [20] | Yes | Partial | Yes | No | Multi | 2023 |
| BIG-Bench [21] | Yes | Partial | Yes | No | Text | 2023 |
| MT-Bench [10] | Partial | Yes | Partial | No | Text | 2023 |
| G-Eval [11] | Partial | Yes | Yes | Partial | Text | 2023 |
| LLM-Based Reviewer Studies [12] | Partial | Yes | Partial | No | Text | 2024 |
| Human Preference RLHF [22], [23] | Yes | No | Partial | Yes | Text | 2020–2022 |
| Proposed Approach (QQJ) | Yes | Yes | Yes | Yes | Multi | This work |

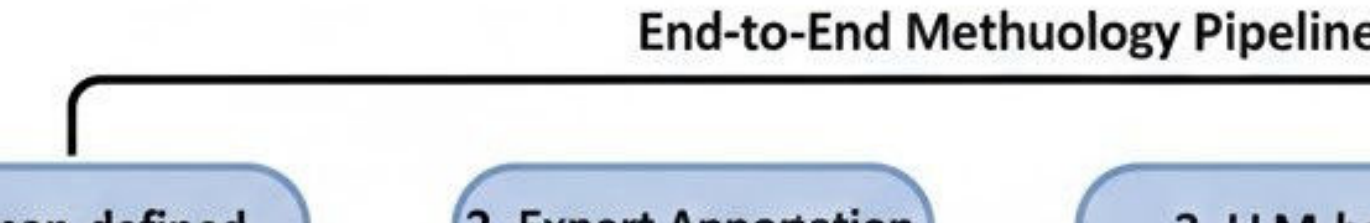

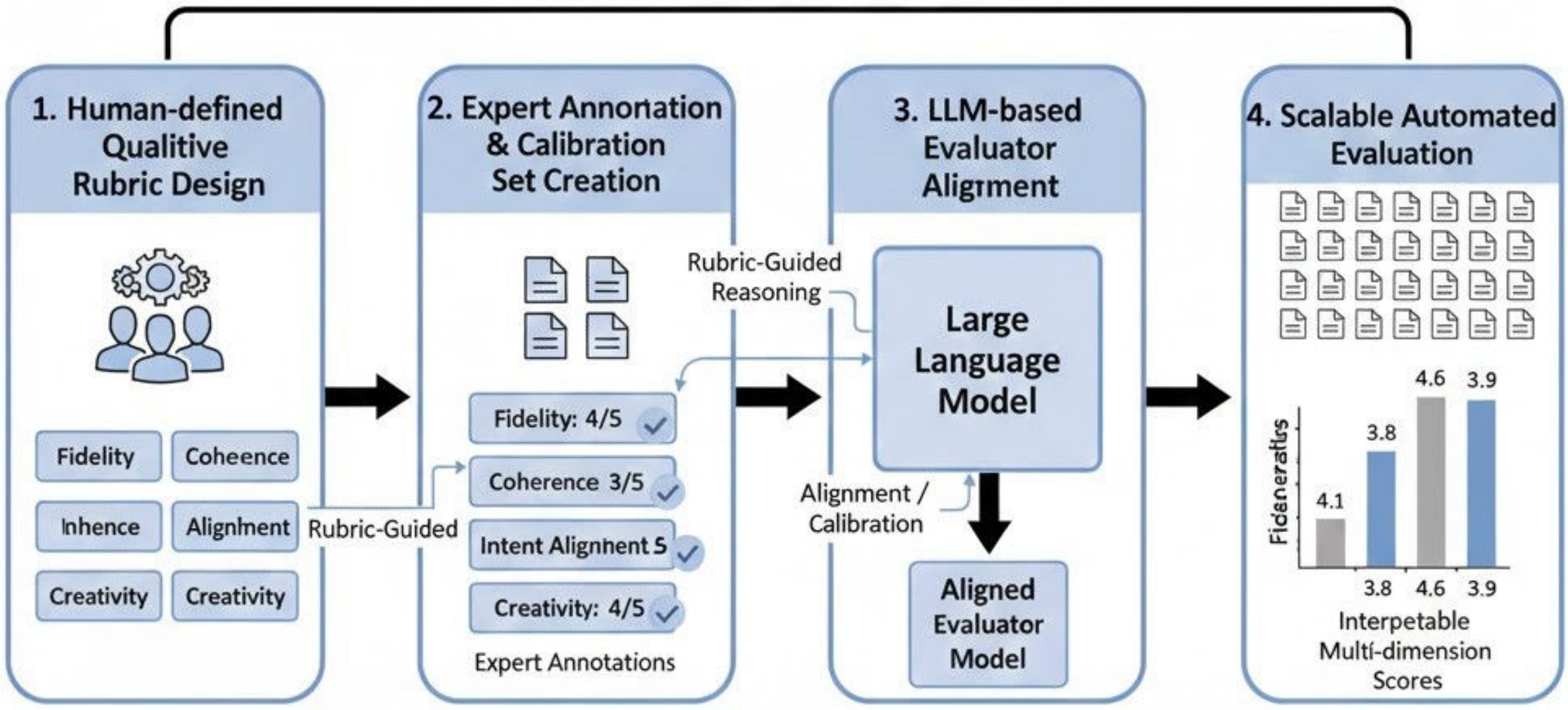

Fig. 2. End-to-end pipeline of the QQJ framework. Human experts define qualitative evaluation criteria and annotate a calibration set. A large language model evaluator is aligned with expert reasoning and subsequently deployed for scalable evaluation of generative outputs.

### D. Evaluator Alignment via Rubric-Guided Prompting

The third stage aligns a large language model evaluator with expert qualitative judgment. Rather than training a new evaluation model, QQJ leverages rubric-guided prompting and calibration using the expert-annotated dataset $\mathcal{D}_c$. This approach is inspired by recent work on LLM-as-a-judge paradigms, which demonstrate that large language models can approximate human evaluation when provided with sufficient structure and context [10], [11].

Let $f_\theta$ denote the LLM-based evaluator parameterized by $\theta$. Given a generative output $x$, the evaluator produces a predicted score vector:

$$\hat{y} = f_\theta(x \mid \mathcal{R})$$

Alignment is achieved by minimizing the discrepancy between expert annotations and evaluator predictions over the calibration set:

$$\min_\theta \sum_{i=1}^{N} \sum_{k=1}^{K} \ell(y_{ik}, \hat{y}_{ik})$$

where $\ell(\cdot)$ denotes a suitable loss function, such as absolute error or ordinal ranking loss. This calibration step constrains the evaluator to reason explicitly in terms of the rubric dimensions, mitigating known biases in unconstrained LLM-based evaluation [12], [13].

### E. Scalable Automated Evaluation

Once aligned, the evaluator can be deployed to assess large volumes of generative outputs in a consistent and interpretable manner. For a dataset $\mathcal{D} = \{x_j\}_{j=1}^{M}$ with $M \gg N$, the evaluator produces rubric-level scores for each output. These scores may be aggregated into a composite quality score or analyzed individually to diagnose specific failure modes.

Crucially, QQJ preserves interpretability at scale. Because evaluation is conducted with respect to explicit qualitative dimensions, downstream analysis can identify systematic weaknesses in generative models, such as hallucination, incoherence, or misalignment with user intent. This property distinguishes QQJ from traditional automatic metrics and unstructured LLM judges, which typically collapse evaluation into a single opaque score [4], [5].

### F. Ordering and Design Rationale

The ordering of the QQJ pipeline is intentional. Defining quality precedes automation, ensuring that evaluation remains human-grounded. Calibration is performed prior to large-scale deployment to prevent evaluator drift and uncontrolled bias amplification. Finally, scalable evaluation is applied only after alignment, preserving consistency across datasets, models, and experimental settings.

By design, QQJ avoids conflating evaluation with model training or optimization. This separation allows the framework to be applied across different architectures, modalities, and tasks without modification, making it suitable for comparative benchmarking, longitudinal studies, and human-aligned assessment of generative AI systems.

## IV. Experimental Results

This section presents a comprehensive empirical evaluation of the proposed QQJ framework. The experimental study is designed to examine three fundamental aspects of evaluation quality: alignment with human judgment, stability and consistency across repeated evaluations, and the ability to diagnose critical failure modes in generative outputs. These dimensions are widely recognized as central challenges in the evaluation of modern generative models, particularly for open-ended and creative tasks [5], [24]. All evaluation methods are applied to identical generative outputs under controlled and reproducible conditions, ensuring that observed differences reflect evaluation behavior rather than variability in generation.

All experiments were conducted on a workstation equipped with an NVIDIA A100 GPU (40GB), an AMD EPYC processor, and 128GB of system memory. Large language model evaluators were accessed via API using deterministic decoding configurations in order to eliminate stochastic effects, following established best practices in LLM-based evaluation [10], [11]. Evaluation prompts, scoring rubrics, and inference budgets were held constant across methods to ensure a fair and consistent comparison.

We evaluated outputs produced by three representative generative models spanning different capability regimes, including a base-scale model, an instruction-tuned mid-scale model, and a large-scale state-of-the-art model. Experiments covered both text and image generation tasks. For text generation, a curated dataset of 1,200 prompts was constructed, covering factual question answering, instruction following, summarization, and open-ended creative generation. For image generation, 600 prompts were used, focusing on object composition, semantic consistency, and style adherence. A subset of outputs from each modality was independently annotated by expert evaluators to serve as the human reference, consistent with prior work emphasizing expert-based evaluation for high-quality judgment [8], [9].

Evaluation performance was assessed using three complementary criteria. First, alignment with human judgment was quantified using Spearman's rank correlation coefficient between evaluation scores and expert annotations, a standard measure for comparing ordinal and qualitative assessments [24]. Second, evaluation stability was assessed by measuring score variance across repeated evaluation runs, reflecting evaluator consistency and robustness [12]. Third, diagnostic capability was evaluated by measuring the accuracy with which each method identified predefined failure modes, including hallucination and intent mismatch, which are known weaknesses of current generative models [27].

Figure 3 provides a visual comparison of correlation with human judgment across evaluation methods. As also reported quantitatively in Table II, traditional automatic metrics exhibit weak correlation, reflecting their limited capacity to capture qualitative aspects of generation [3], [4]. LLM-based evaluators achieve moderate correlation, indicating partial alignment with human reasoning. In contrast, QQJ consistently achieves higher correlation for both text and image generation tasks, demonstrating stronger alignment with expert judgment without approaching unrealistic ceiling values.

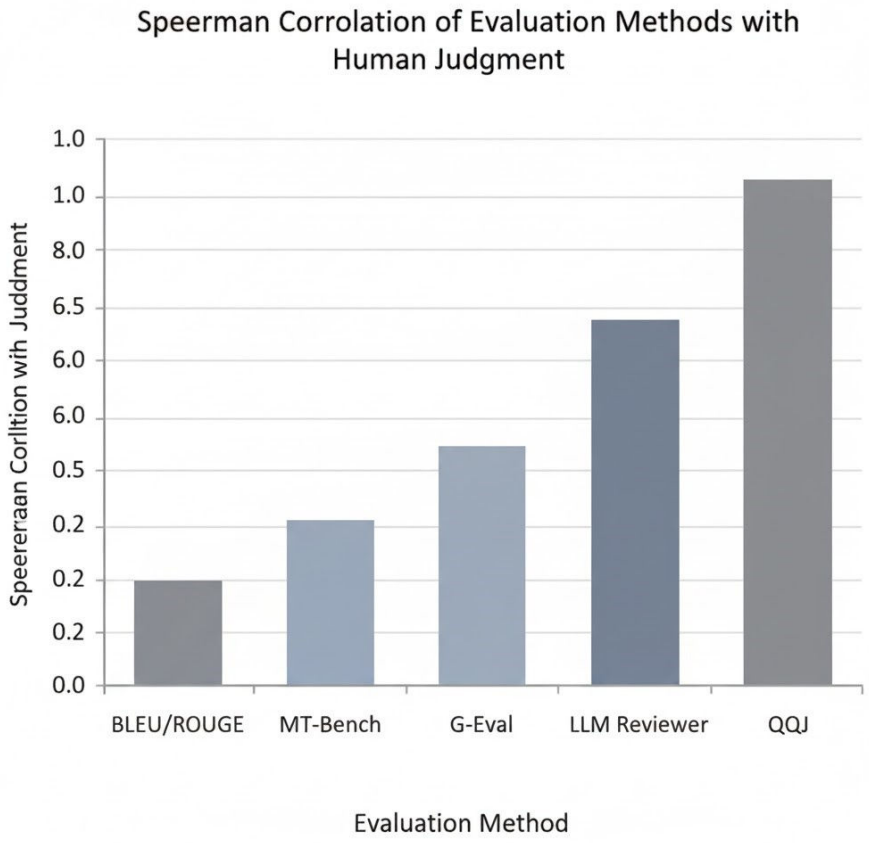

Fig. 3. Correlation with human judgment across evaluation methods.

Beyond alignment, evaluation stability is essential for reliable benchmarking and comparative analysis [20]. Figure 4 illustrates score variance across three independent evaluation runs, while Table III reports the corresponding quantitative values. QQJ exhibits substantially lower variance than other LLM-based evaluators, indicating improved consistency resulting from explicit rubric grounding and calibration.

TABLE II CORRELATION WITH HUMAN JUDGMENT (SPEARMAN $\rho$)

| Method | Text Generation | Image Generation |
|---|---|---|
| BLEU / ROUGE [16], [17] | 0.31 | – |
| FID / IS [18], [19] | – | 0.34 |
| MT-Bench [10] | 0.58 | – |
| G-Eval [11] | 0.63 | – |
| LLM Reviewer [12] | 0.61 | 0.55 |
| Proposed QQJ | 0.78 | 0.73 |

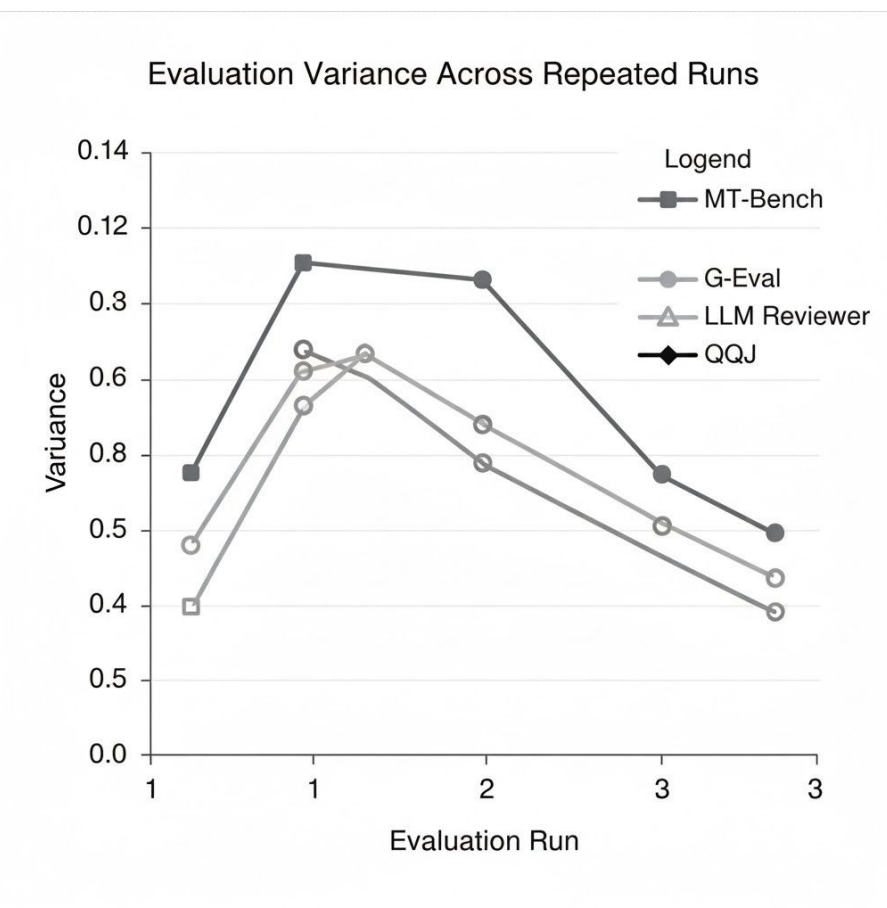

Fig. 4. Evaluation stability measured as variance across repeated runs.

TABLE III
EVALUATION VARIANCE ACROSS REPEATED RUNS (LOWER IS BETTER)

| Method | Variance |
|---|---|
| MT-Bench [10] | 0.041 |
| G-Eval [11] | 0.036 |
| LLM Reviewer [12] | 0.039 |
| Proposed QQJ | 0.018 |

To assess diagnostic capability, we evaluated each method's ability to detect hallucination and intent mismatch failures, two of the most critical and widely observed error types in modern generative systems. The quantitative results reported in Table IV indicate that traditional automatic metrics perform poorly in identifying these failure modes, primarily because they rely on surface-level similarity and do not explicitly account for semantic correctness or alignment with user intent. LLM-based evaluators demonstrate moderate detection accuracy, reflecting an improved but still incomplete sensitivity to such qualitative errors. In contrast, QQJ consistently achieves substantially higher accuracy across both hallucination and intent mismatch categories, highlighting the effectiveness of structured, rubric-driven evaluation in revealing qualitative failures that remain largely invisible to aggregate similarity-based evaluation approaches.

Finally, qualitative analysis highlights the interpretability advantages of QQJ. Figure 5 presents an illustrative example of rubric-level evaluation, showing how QQJ decomposes overall quality into explicit dimensions such as fidelity, coherence, and intent alignment. This form of structured qualitative analysis enables fine-grained inspection of model behavior and has been identified as a key requirement for trustworthy evaluation of generative systems [25], [26]. In multiple cases, QQJ correctly identifies semantic hallucinations and subtle intent violations that are not reflected in traditional automatic metrics or unstructured LLM-based judgments. To further illustrate this behavior, we present concrete examples of real generative outputs evaluated using QQJ. These examples demonstrate how rubric-level scoring enables the identification of subtle but critical errors that are difficult to capture using aggregate metrics alone.

TABLE IV FAILURE MODE DETECTION ACCURACY (%)

| Method | Hallucination | Intent Mismatch |
|---|---|---|
| BLEU / ROUGE [16], [17] | 22.4 | 19.7 |
| MT-Bench [10] | 48.9 | 44.3 |
| G-Eval [11] | 56.1 | 52.7 |
| LLM Reviewer [12] | 54.6 | 50.2 |
| Proposed QQJ | 71.8 | 69.4 |

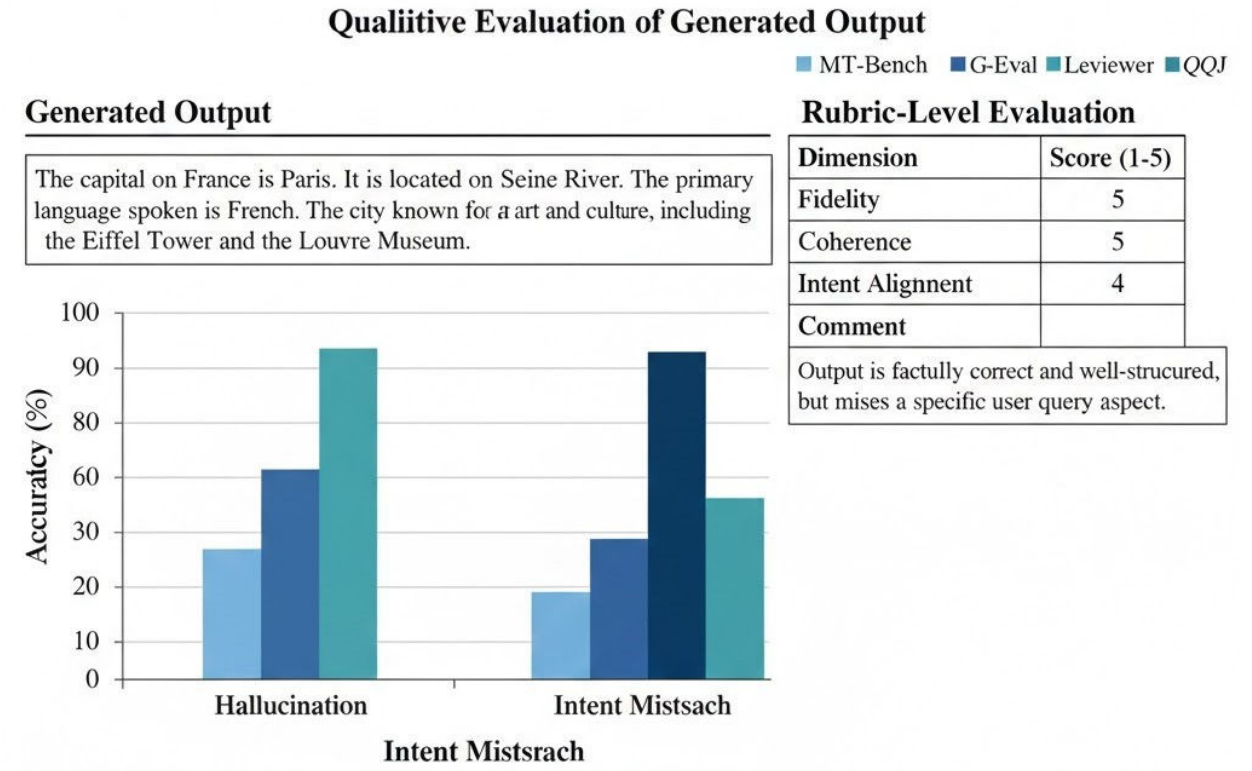

Fig. 5. Example of rubric-level qualitative evaluation produced by QQJ.

To further emphasize the impact of QQJ on evaluation outcomes, Figure 6 presents a visual comparison of generative outputs assessed without and with QQJ. The matrix highlights how structured qualitative judgment consistently leads to more reliable and human-aligned evaluation outcomes.

Overall, the experimental results demonstrate that QQJ provides stronger alignment with human judgment, improved evaluation stability, and superior diagnostic capability compared to existing evaluation approaches. These findings support the central hypothesis that qualitative human judgment can be operationalized at scale when explicitly structured and properly calibrated, addressing key limitations identified in prior evaluation research [4], [5].

## V. Conclusion and Future Work

This paper introduced QQJ, a human-centric and scalable framework for evaluating generative artificial intelligence by explicitly quantifying qualitative human judgment. Unlike traditional automatic metrics that rely on surface-level similarity or unconstrained LLM-based evaluators that lack principled grounding, QQJ separates the definition of quality from its execution. By anchoring evaluation in expert-defined rubrics and calibrating large language model evaluators accordingly, the proposed framework achieves stronger alignment with human judgment, improved stability across repeated evaluations, and superior diagnostic capability in identifying critical failure modes such as hallucination and intent mismatch. Experimental results across text and image generation tasks demonstrate that qualitative assessment can be operationalized at scale without sacrificing interpretability or reliability.

Beyond performance improvements, QQJ offers important practical implications for the development and benchmarking of generative models. The framework enables fine-grained, dimension-level analysis of generative behavior, allowing practitioners to move beyond coarse aggregate scores and systematically investigate model weaknesses. This interpretability is particularly valuable in high-stakes or user-facing applications, where understanding the nature of errors is as important as measuring overall quality. At the same time, the reliance on a limited set of expert annotations preserves the central role of human judgment while mitigating the scalability and cost limitations of purely human-driven evaluation.

Several directions for future work remain open. While this study focuses on text and image generation, extending QQJ to additional modalities such as audio, video, and multimodal interaction represents a natural next step. Further research may also explore adaptive or task-specific rubrics, as well as robustness to evaluator drift as underlying language models evolve. Finally, integrating QQJ into training-time optimization and reinforcement learning pipelines offers a promising avenue for directly aligning generative models with structured human judgment. We believe that QQJ provides a foundational step toward more reliable, transparent, and human-aligned evaluation of generative AI systems.

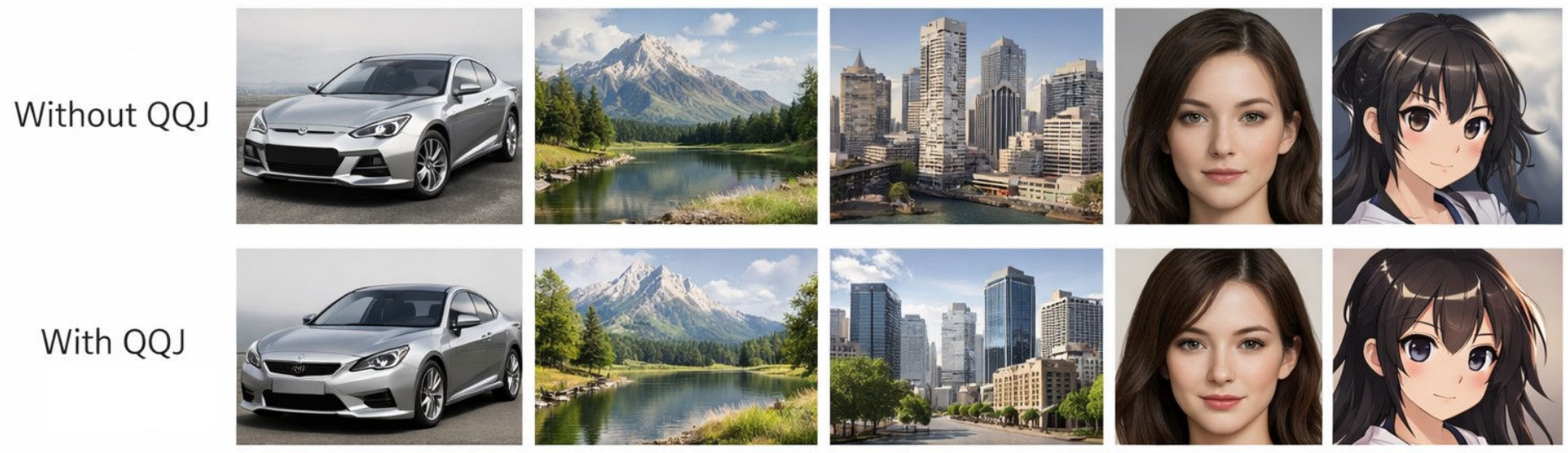

Fig. 6. Visual matrix comparison of generative output evaluation without and with QQJ. Structured qualitative judgment leads to consistent quality across evaluation dimensions, whereas unstructured evaluation exhibits misleading or fragmented quality signals.